\documentclass{article}
\usepackage{graphicx} 
\usepackage{amssymb}
\title{Retrieval of phonemes and Kohonen algorithm}
\author{Orchidea Maria Lecian\\ \small Sapienza University of Rome,\\ \small Department of Civil and Industrial Engineering,\\ \small Via Eudossiana, 18- 00184 Rome, Italy;\\ \small Sapienza University of Rome,\\ \small Department of Information Engineering,\\ \small Via Eudossiana, 18- 00184 Rome, Italy;\\ \small orchideamaria.lecian@uniroma1.it\\ \normalsize
Brunello Tirozzi\\
\small
Sapienza Unviersity of Rome,\\ \small Physics Department,\\ \small Piazzale Aldo Mororo, 5- 00185 Rome, Italy;\\ \small brunellotirozzi@gmail.com \normalsize}
\date{June 2023}

\begin{document}

\maketitle
\section*{Key-words} Phoneme-retrieval; voice recognition; vowels recognition.
\section*{Abstract}
A phoneme-retrieval technique is proposed, which is due to the particular way of the construction of the network.
An initial set of neurons is given. The number of these neurons is approximately equal to the number 
of typical structures of the data. For example if the network is built for voice retrieval then the number of neurons
must be equal to the number of characteristic phonemes of the alphabet of the language spoken by the social group 
to which the particular person belongs. Usually this task is very complicated and the network can depend critically on the samples
used for the learning. If the network is built for image retrieval then it works only if the data to be retrieved belong to a particular set
of images. If the network is built for voice recognition it works only for some particular set of words. A typical example is the
words used for the flight of airplanes. For example a command like the "airplane should make a turn of 120 degrees towards the east"
can be easily recognized by the network if a suitable learning procedure is used.

\section{Introduction\label{section1}}
The phonemes are the fundamental elements of a spoken language. Vowels and consonants are two particular phonemes, and they are produced in different mechanisms.\\
A vowel is generated after the use of the vocal cords, which give rise to a periodic acoustic signal, which is qualified after precise spectral components. The differences among vowels are due to te articulations and to the opening of the lips and to that of the jaw: the corresponding signal does not exhibit random components nor disturbance ones.\\
On the contrary, the production of consonants does not involve vocal cords, since it is due to a constriction of the mouth, where the latter, on its turn, induces a turbulence on the air which blows out of the lungs. The turbulence creates a random component in the vocal signal, or a combination of noise and periodic signal. There are also consonants which are due to the mouth and to the nose.\\
The spectrum of the vowels demonstrates resonances which are a multiple of the fundamental frequency,which coincides with the frequency of the oscillations of the vocal cords; the pertinent power spectrum exhibits maxima in the correspondence of the multiples of the fundamental frequencies.
\section{The power spectrum}
The power spectrum is the quantity after which vowels and consonants are parameterised.\\
{\bf Definition 1}.Let $x(t)$ be a stochastic process  defined on a space of probability $(\Omega, P, \Sigma)$; the power spectrum $S(\omega)$ is defined after the relation
\begin{equation}
    S(\lambda)= E \mid \int e^{i\lambda t} x(t) dt \mid^2
\end{equation}
where $E$ is the expectation value with respect to the probability $P$.\\
Vowels are characterised after a concentrated spectrum on the lower part of the spectrum and on the medium one, i.e. for frequencies smaller than $3$ $K$ $Hz$.\\
The fundamental frequencies of the vowels can be easily identified, as the signal is periodic, and the noise component is small. The vowels differ one from the other for the position of the fundamental frequencies; in particular, it is enough to consider the first one and the second one. The form of the spectrum of a phoneme depends also on the word which contains it and on the pronunciation of the speaker.\\
The vector that represents a phoneme is built staring from the power spectrum $S(\lambda)$. Since the samples of the spoken language are obtained after measures at discrete times, it is necessary to perform a Fourier transform of a sequence $x(t_k)$. It is therefore more apt to use the fast Fourier transform (FFT) \cite{ref19}. The algorithm consists in dividing the data $x(k)\equiv x(t_k)$, $k=1, ..., 2K$ into two subsets, i.e. one consisting of the data of even index, and the other of the data of odd index; a set of complex data $z(k)$ with index $k=1, ..., K$ is introduced thereafter, such that the real part $Re[z(k)]$ equals the first $K$ data with $k=1, ..., K$, and the imaginary part of $z(k)$, $Im[z(k)]$ to the second part.\\
It is easy to find the relation which colligates the Fourier transform of the reals part of $z(k)$ with that of the imaginary part of $z(k)$, such that the problem is reconducted to the calculation of the Fourier transform for $K$ data only, instead of one for the $2K$ data which were originally given.\\
If $K=2^N$, it is straightforward to verify that the number of operations $K^2$ reduces to $K log_2K=2^NN$ after the iteration of the procedure. The interval of frequencies in which the power spectrum is defined is divided in a certain number of intervals (or 'band'). These intervals correspond to the ways the human ears work at the variation of the frequencies. The pattern $x$ consists of a vector that has as many components as the number of the bands, and the value f the $i-th$ component equals the mean value of the power spectrum of the $i-th$ band. The sampling of the signal has to be accomplished at a proper frequency which avoids the distortion of the signal, after the appropriate theorem \cite{ref23}.\\
It is the purpose of the present paper to analyse the case of vowels, as it is easier to isolate the stationary part of the acoustic signal which corresponds to the phoneme, as consonants give raise to a signal which exhibits the presence of a strong noise.\\
The data $x(t)$ which measure the acoustic signal are grouped in boxes of $512$, and the FFT is applied to every box. The blocks are issued from the first datum, then from the second one, and so on. The FFT is averaged among the blocks which correspond to the same part of the signal; in other words, the Fourier transforms are averaged, among the blocks which correspond to the blocks which are part of the same phoneme. The average is accomplished because the division zone between one phoneme and another is not so easily outlined. As a further problem to be solved, it is worth mentioning that the Fourier transform accomplished only on the block containing $512$ data is not the true Fourier transform, as the integral defining it is an integral on $\mathbb{R}$. There exists a theory \cite{ref23} which allows one to correct this miscalculation, according to which it is necessary to multiply the succession of the data which is to be considered in the summation (or in the integral) times a certain function which depends on the shape of the data block (which is called a 'time window'). There exist several functions: the Welch functions, the Parzen function, the Hanning function, and so on: it is customary to verify how the pattern vecor depends on the choice of these functions. Of course, this correction has to be accomplished before the calculation of the power spectrum. In \cite{ref21} the verification is presented, of the issue that the vowels $a, o, e, i, u$, extracted from a certain succession of words, in the application of a rectangular window, i.e. the multiplication time the characteristic function of the block constituted of $512$ and the multiplication times the Hanning function do not lead to very different results as far as the frequencies less than $3$ $Hz$, which is the interval among which the power spectrum of the spoken language is concentrated.\\
To summarise, the patterns are extracted from the vocal signal after the following operations:
\begin{itemize}
    \item 1) For each group of $512$ data, the FFT is calculated with the above-mentioned corrections, and the power spectrum is determined;
    \item 2) the frequency interval $0-5000$ $Hz$ is divided into $15$ intervals (or 'channels'): from $200$ to $3000$ $Hz$, $12$ intervals are considered, of breadth $233$ $Hz$, while from $3000$ $Hz$ to $5000$ $Hz$ only $3$ intervals are considered, of breadth $667$ $Hz$;
    \item 3) in each channel, the average of the power spectrum is calculated;
    \item the vector $\vec{x}$, which consists of $15$ components, constructed this way is normalised to $1$ in the Euclidean norm, for a special convergence theorem of the weights to be applied, in a particular Kohonen network \cite{sectionV}.
\end{itemize}
The described construction can be now applied. Given a phoneme, which is represented after a vector built in the appropriate manner, the different position of the phoneme in the different words and the different pronunciation due to the inflection of the voice allow for the existence of a set of vector $A(\vec{x})$ which correspond to that phoneme. It is obvious that the phonemes generate a Voronoi partition, and the Kohonen partition algorithm should allow one to construct such a partition, together with the vectors $\vec{x}_i$ which define the partition. The dynamics of the winning neuron is applied to prove to theorem of convergence of the weights. Nevertheless, the theorem which is aimed to be proven is based on a non-linear dynamics of the weights of the network. In \cite{ref21}, a validation of this version of the theorem was provided with; nevertheless, a recognition of vowels only was achieved only in $51\% $ of cases. The reason of the inefficient performance was outlined in that the phonemes cannot be restricted to vowels an consonants only, but the transitions between phonemes have to be introduced, as in \cite{ref22}. In \cite{ref22}, more sensitive parameters were introduced as well, as far as the power spectrum is concerned, which lead the patter vector to consist of $500$ components.
\section{A particular Kohonen algorithm\label{section2}}
The present section is aimed at discussing a network consisting of $n$ input neurons and of $n$ output neurons.. At each knot of the first type the same pattern $\vec{x}\in \mathbb{R}^n$ is presented, which represent a particular phoneme. A weight vector $m_i\in\mathbb{R}^n$ is associated with the $i-th$ output neuron $i=1, ..., N$. Each input neuron is connected with all the output neurons. The weights $m_i(t)$ satisfy the $n-$dimensional Riccati equation
\begin{equation}\label{v61}
\dot{m}_i=\alpha x(t) -\beta m_i\sum_{j=1}^{j=N}m_{ij}x_j(t), \ \ \alpha, \beta>0
\end{equation}
The dynamics was introduced by Kohonen to take into account the non-linear response of the circuits which can eventually realise this algorithm on some computer. As soon as the dynamics is opportunely discretised, the dynamics is applied on the winning euro only, i.e. on the neuron $i$ whose weight is as closest as possible to the input vector $x$ i the Euclidean distance. The evolution equation of this dynamics is non-linear. The vector which corresponds to a pattern is constructed as described in Section \ref{section1}. It is possible to state that, if an input time sequence $x(t)$, issued from the spoken language of a chosen person, is presented to the network, a structure of vectors $\hat{x}_k$ should be obtained, where the latter define the Voronoi partition associated with the set of Phonemes generated by the chosen person. It is expected that two different persons give rise to two different partitions.  For it to be accomplished, a convergence and stability Voronoi is necessary. As the dynamics described after Eq. (\ref{v61}) is non-linear, the convergence theorem in probability is substituted by a Voronoi which states the stability of the weights $m(t)$ in the asymptotic limit when the vector $x(t)$ varies within a neighbourhood which is small enough. The voice recognition after this kind of network has not been applied successfully yet,a d, up to now, there are programs which are able to recognise only a limited number of words, if applied to one person only, after a sufficiently-enough long instruction time, within a certain error. There exist also other algorithms for the voice recognition \cite{ref24}, \cite{ref25}.\\
The vector $\hat{x}_k$ which generates the atom $A(\hat{x}_k)$ of the Voronoi partition is the vector with the least distance from all the other vectors of $A(\hat{x}_k)$, which is named the central vector.\\
The characteristic radius $r_k$ of the atom $A(\hat{x}_k)$ is the maximum distance between the central vector and all the other vectors belonging to $A(\hat{x}_k)$.\\
The distance $\delta_{kl}$ between the central vectors of the atoms $A(\hat{x}_k)$ and $A(\hat{x}_l)$ and the minimum distance between the atoms is defined:
\begin{equation}\label{v62}
    \delta=min_{k, l} \delta_{kl}.
\end{equation}
{\bf Definition 2:} {\it If an unknown pattern $x$ is presented to the network, the neuron $i$ is found, such that its weight vector is endowed with the minimum distance $\rho$ from the input $x$. Let $A(\hat{x}_k)$ be the atom of the partition to which the weights of the neuron $i$ belong: if $\rho\le r_k$, then the pattern $x$ is recognised as the $k-$phoneme; differently, it is not recognised.}\\
For a learning process to give rise to weights similar to the central vectors of each phoneme or atom of the Voronoi partition, it is necessary to prove the stability of the Riccati equation with respect to the variation of the input function $x(t)$. More precisely, two theorems need to be demonstrated \cite{ref20}.\\
{\bf Theorem 1}: If in Eq. (\ref{v61})a constant function $\hat{x}_k$ is introduced, the the limit of the solution is proportional to $\hat{x}_k$, and the vector $m(t)$ approaches to this value with exponential velocity.\\
As it happens during the instruction procedure of the network $x(t)=\hat{x}_k+y(t)$, on the contrary, one has the following\\
{\bf Theorem 2}: If the norm of the perturbation $y(t)$ is small, the norm of the variation of the solution of the equation of the evolution of the weights is minorised by a constant multiplied times this small constant.\\
In other words, the Riccati equation is stable with respect to variations of the input vector. This property allows one for the construction of the Voronoi partition if the learning algorithm here chosen is used. This results remains valid also if perturbation $y(t)$ is a stochastic process of continuous trajectories; the almost-everywhere convergence is not obtained any more from the learning dynamics: only stability is achieved. It is important to remark that stability holds only if the components of the input vector are all strictly greater than or equal to a fixed positive number $\gamma$; this behaviour was noticed also by Kohonen \cite{ref26}, but no satisfying explanation was given thereafter. This hypothesis provides one also with the motivation of the choice of the power spectrum as a representation of the voice.\\
{\bf Theorem 3}:
If $\mid y(t)\mid<\delta$, $\forall t$, then there is stability if
\begin{equation}
\delta<\gamma/8;
\end{equation}
it is important that this criterion be satisfied when $\delta$ coincides with the value given by Eq. (\ref{v62}). In \cite{ref21} work has been developed to prove Theorem 3. As a sample, $44$ words pronounced by a woman were analysed. The patterns pertinent to the vowels $a, o, e, i, u$ were extracted. $33$ samples of $a$ were obtained; $33$ samples of $o$ were obtained; $20$ samples of $e$ were obtained; $13$ samples of $i$ were obtained. Because only $3$ samples of $u$ were obtained, the latter vowel was excluded form the study. The constant $\gamma$ is of order $0.9\cdot10^{-4}$. The convergence of the Riccati equation was obtained after a numerical integration of $150$ steps, and the value was differing from that forecast by the theoretical one of a quantity smaller than $3\cdot10^{-7}$, in the case of a constant input. Furthermore, the numerical solution confirms that the approach to the limiting value is an exponential one. The stability condition was not matched because the value of $\delta$ was calculated as $\delta<\gamma/8$; nevertheless, after the numerical integration of Eq. (\ref{v61}), this property was verified by means of the data available.\\
\section{Discussion}
It is now appropriate to state the following Definitions and Theorems.\\
{\bf Definition 3}: let $x(t)$ be the input value of the network, and it is the same for each of the $N$ neurons which constitute it; a weight vector $m_i(t)\in \mathbb{R}^n$ is associated with each neuron $i$; the output of a generic neuron is given by 
\begin{equation}\nonumber \eta=(m, x)
\end{equation} where $( , )$ is the Euclidean scalar $n-$dimensional product.\\
The output of the network given in this definition corresponds to what learnt from the recognition; indeed, the condition according to which the Euclidean distance between the weight of the winning neuron and the input pattern be minimal corresponds to the fact that $\eta$ is the maximum: it is therefore appropriate to state that, during the recognition, only the neuron whose maximum output, according to the previous definition, be active.\\
{\bf Definition 4}: The phonemes form a set of $n-$ dimensional constant vectors $\hat{x}^1, ..., \hat{x}^M$, $M\le N$, such that
\begin{equation}\label{v63}
\mid \mid \hat{x}^i\mid \mid =1, \ \ \mid \mid \hat{x}^i-\hat{x}^j\mid\mid>\delta
\end{equation}
for $i, j=1, ..., M$, where $\delta$ is the parameter previously defined, and $\mid\mid\cdot\mid\mid$ is the Euclidean norm of the space $\mathbb{R}^n$.\\
This definition equals the statement that the central vectors of the Voronoi partition are normalised to $1$, and that there exists a minimum distance between them, which is an important parameter within the construction.\\
{\bf Definition 5}: The 'perturbed' set of the phonemes is a vector function
\begin{equation}\label{v64}
x^i(t)=\hat{x}^i+y^i(t)    
\end{equation}
with $\mid\mid y^i(t)\mid\mid^2\le\delta^2$ for $i=1, ..., M$ and $y^i(t)$ continuous.\\
In this definition, the fact is established, that the set of phonemes by which the instruction of the network is done is given by the vector which really represents the phoneme, to which a quantity is added, which can be also random as far as it is continuous, which represents all the random fluctuations due to the accent of a person, to the position of the phoneme in the word, and so on. It is remarked that the construction is meaningful if this perturbation is smaller than $\delta$; differently, there is a superposition between the data which instruct the network.\\
{\bf Definition 6}: The evolution equation of the weight vector of the generic neuron $i$ is as follows:
\begin{equation}\label{v65}
\dot{m}_i=\alpha x^k(t)-\beta m_{ij}x_j^k(t),
\end{equation}
where $x^t(k)$ is the $k-th$ perturbed phoneme, $\alpha$ and $\beta$ are positive constants, which are fixed, which depend on the characteristics of the particular circuit which realises the neural circuit.\\
It is interesting to remark that, if one takes $x(t)=\hat{x}^k(t)$, with $k$ fixed, the the $m^*$ chosen as
\begin{equation}
    m^+=\sqrt{\frac{\alpha}{\beta}}\hat{x}^k(t)
\end{equation}
is a fixed point of Eq. (\ref{v65}). Let $m^0(t)$ the solution of Eq. (\ref{v65})
 for $x(t)=\hat{x}^k(t)$. There holds the following\\
 {\bf Theorem 4}: For each initial condition $m(0)$, one has
 \begin{equation}\label{v67}
     m^0=\hat{x}^k(t)\sqrt{\frac{\alpha}{\beta}}+O(-\sqrt{\alpha\beta}t).
 \end{equation}
 Let $m(t)$ be the solution of the evolution equation for the weights $x(t)=\hat{x}^k(t)$, and let $v$, $v=m-m^0$, the variation of the solution with respect to the solution $m^0$. The following Definition 7 and the following Theorem 5 are equivalent to stating the previous discussion, which demonstrates that the instruction of the neural network is meaningful only if the fluctuations which are present in the set of instruction of the network do not let the central vector of the Voronoi partition, which the instruction process builds, vary much.\\
 {\bf Definition 7}: The network formed by the $N$ neurons and their weights is stable with respect to the variation of the input $\hat{x}^k(t)$ if it is possible to find $\delta$ such that, for each $y(t)$ continuous, $\mid\mid y(t)\mid\mid\le\delta$, the exists $c$ such that
 \begin{equation}
 \mid\mid v(t)\mid\mid\le C\delta,
 \end{equation}
 where $v$ is the variation of the previously-defined solution.\\
 It is now possible to expose the theorem which states the stability of the system in the sense of Definition 7. The theorem is valid also if the perturbation is a random function, as the only property which is requested is the continuity which is verified for many stochastic processes present in  nature.\\
 {\bf Theorem 5}: Let $x^k(t)=\hat{x}+y^k(t)$ a continuous vector function, and let $\gamma>0$ such that\begin{equation}\label{v68}
 \hat{x}^k_i\ge\gamma, \ \ \forall i=1, ..., n
 \end{equation}
 and $m_i(0)>0$: then one has
 \begin{equation}\label{v69}
     \delta<\frac{\gamma}{8}
 \end{equation}
 to determine $T=T(y^k)$ such that
 \begin{equation}\label{v610}
 \mid v_i(t)\le C\delta, \ \ for\ \  t\ge T(y^k), i=1, ...,n
 \end{equation}
 \begin{equation}
     C=\frac{16}{\gamma}\sqrt{\frac{\alpha}{\beta}}.
 \end{equation}
 \section{Outlook}
 Self-organising maps have been tested for the recognition of word boundaries in \cite{refa}.\
 Coding strategies between layers are discussed in 
 \cite{refb}.\\
 Self-organising neural networks are analsyed in \cite{refc} as far as the validity of the technique to span the speech space.\\
 The ability of time-dependent self-organising maps can be used to determine the time-dependent features of the input speech signal \cite{refd} .\\ 
 The consequences of the modifications of the input signal are studied in \cite{refe}, \cite{reff}.\\
 The analysis of consonants has been scrutinised from different techniques; as a main result, the analysis of consonants is dependent of the chosen language \cite{refg}, \cite{wss89}, \cite{gcy}.\\
 The dynamic stability of the neural networks has been investigated in \cite{mtt92}.\\ 
 The Kohonen dynamics in a dynamically expanding context has been considered in \cite{tko1988}.\\
 An example of winner-take-all neural network is given in \cite{fck96}.
 
 \end{document}